%% file: sample.tex
\documentclass[twoside,11pt]{article}

\usepackage{blindtext}

\usepackage{jmlr2e}

\usepackage{lastpage}

\ShortHeadings{Causal-learn: Causal Discovery in Python}{Zheng et al.}
\firstpageno{1}

\usepackage{listings}
\usepackage{xcolor}  

\begin{document}

\title{Causal-learn: Causal Discovery in Python}

\author{%
\name Yujia Zheng$^{1, 6}$ \email yujiazh@cmu.edu \\
\name Biwei Huang$^{2}$ \email bih007@ucsd.edu \\
\name Wei Chen$^{3}$ \email chenweidelight@gmail.com \\
\name Joseph Ramsey$^{1}$ \email jdramsey@andrew.cmu.edu \\
\name Mingming Gong$^{4}$ \email mingming.gong@unimelb.edu.au \\
\name Ruichu Cai$^{3}$ \email cairuichu@gmail.com \\
\name Shohei Shimizu$^{5,7}$ \email shohei-shimizu@biwako.shiga-u.ac.jp \\
\name Peter Spirtes$^{1}$ \email ps7z@andrew.cmu.edu \\
\name Kun Zhang$^{1,6}$ \email kunz1@cmu.edu \\
\addr $^{1}$ Carnegie Mellon University\\
\addr $^{2}$ University of California, San Diego\\
\addr $^{3}$ Guangdong University of Technology\\
\addr $^{4}$ University of Melbourne\\
\addr $^{5}$ Shiga University\\
\addr $^{6}$ Mohamed bin Zayed University of Artificial Intelligence\\
\addr $^{7}$ RIKEN
}



\maketitle

\begin{abstract}
Causal discovery aims at revealing causal relations from observational data, which is a fundamental task in science and engineering. We describe \textit{causal-learn}, an open-source Python library for causal discovery. This library focuses on bringing a comprehensive collection of causal discovery methods to both practitioners and researchers. It provides easy-to-use APIs for non-specialists, modular building blocks for developers, detailed documentation for learners, and comprehensive methods for all. Different from previous packages in R or Java, \textit{causal-learn} is fully developed in Python, which could be more in tune with the recent preference shift in programming languages within related communities. The library is available at \url{https://github.com/py-why/causal-learn}.
\end{abstract}


\input{sections/introduction}
\input{sections/design}

\input{sections/example}

\input{sections/conclusion}

\acks{We are grateful for the collective efforts of all open-source contributors that continue to foster the growth of causal-learn. Especially, we would like to thank Yuequn Liu, Zhiyi Huang, Feng Xie, Haoyue Dai, Erdun Gao, Aoqi Zuo, Takashi Nicholas Maeda, Takashi Ikeuchi, Madelyn Glymour, Ruibo Tu, Wai-Yin Lam, Ignavier Ng, Bryan Andrews, Yewen Fan, and Xiangchen Song. The work of MG is supported in part by ARC DE210101624. The work of RC is supported in part by National Key R\&D Program of China (2021ZD0111501). This project is partially supported by the National Institutes of Health (NIH) under Contract R01HL159805, by the NSF-Convergence Accelerator Track-D award \#2134901, by a grant from Apple Inc., a grant from KDDI Research Inc, and generous gifts from Salesforce Inc., Microsoft Research, and Amazon Research.}


\vskip 0.2in
\bibliography{sample}

\end{document}

%% file: sections/introduction.tex
\section{Introduction}
A traditional way to uncover causal relationships is to resort to interventions or randomized experiments, which are often impractical due to their cost or logistical limitations. Hence, the importance of causal discovery, i.e., the process of revealing causal information through the analysis of purely observational data, has become increasingly apparent across diverse disciplines, including genomics, ecology, neuroscience, and epidemiology, among others \citep{glymour2019review}. For instance, in genomics, causal discovery has been instrumental in understanding the relationships between certain genes and diseases. Researchers might not have the resources to manipulate gene expressions, but they can analyze observational data, which are usually widely available, such as genomic databases, to uncover potential causal relationships. This can lead to breakthroughs in disease treatment and prevention strategies without the cost of traditional experimentation.

\looseness=-1
Current strategies for causal discovery can be broadly classified into constraint-based, score-based, functional causal models-based, and methods that recover latent variables. Constraint-based and score-based methods have been employed for causal discovery since the 1990s, using conditional independence relationships in data to uncover information about the underlying causal structure. Algorithms such as Peter-Clark (PC) \citep{spirtes2000causation} and Fast Causal Inference (FCI) \citep{spirtes1995causal} are popular, with PC assuming causal sufficiency and FCI handling latent confounders. In cases without latent confounders, score-based algorithms like the Greedy Equivalence Search (GES) \citep{chickering2002optimal} aim to find the causal structure by optimizing a score function. These methods provide asymptotically correct results, accommodating various data distributions and functional relations but do not necessarily provide complete causal information as they usually output Markov equivalence classes of causal structures (graphs within the same Markov equivalence class have the same conditional independence relations among the variables). 

\looseness=-1
On the other hand, algorithms based on Functional Causal Models (FCMs) have exhibited the ability to distinguish between different Directed Acyclic Graphs (DAGs) within the same equivalence class, thanks to additional assumptions on the data distribution beyond conditional independence relations. An FCM represents the effect variable as a function of the direct causes and a noise term; it renders causal direction identifiable due to the independence condition between the noise and cause: one can show that under appropriate assumptions on the functional model class and distributions of the involved variables, the estimated noise cannot be independent from the hypothetical cause in the reverse direction \citep{shimizu2006linear,hoyer2008nonlinear,zhang2009identifiability}. More recently, the Generalized Independent Noise condition (GIN) \citep{xie2020generalized} has demonstrated its potential in learning hidden causal variables and their relations in the linear, non-Gaussian case.

To equip both practitioners and researchers with computational tools, several packages have been developed for or can be adapted for causal discovery. The Java library TETRAD \citep{glymour1986causal, scheines1998tetrad, ramsey2018tetrad} contains a variety of well-tested causal discovery algorithms and has been continuously developed and maintained for over 40 years; R packages pcalg \citep{pcalg} and bnlearn \citep{bnlearn} also include some classical constraint-based and score-based methods such as PC and GES. However, these tools are based on Java or R, which may not align with the recent trend favoring Python in certain communities, particularly within machine learning. While there are Python wrappers available for these packages (e.g., py-tetrad \citep{pytetrad}/py-causal \citep{pycausal} for TETRAD, and Causal Discovery Toolbox \citep{kalainathan2020causal} for pcalg and bnlearn), they still rely on Java or R. This dependency can complicate deployment and does not cater directly to Python users seeking to develop their own methods based on an existing codebase. Thus, there is a pronounced need for a Python package that covers representative causal discovery algorithms across all primary categories. Such a tool would significantly benefit a diverse range of users by providing access to both classical methods and the latest advancements in causal discovery.

\looseness=-1
In this paper, we describe \textit{causal-learn}, an open-source python library for causal discovery. The library incorporates an extensive range of causal discovery algorithms, providing accessible APIs and thorough documentation to cater to a diversity of practical requirements and data assumptions. Moreover, it provides independent modules for specific functionalities, such as (conditional) independence tests, score functions, graph operations, and evaluation metrics, thereby facilitating custom needs and fostering the development of user-defined methods. An essential attribute of causal-learn is its full implementation in Python, eliminating dependencies on any other programming languages. As such, users are not required to have expertise in Java or R, enhancing the ease of integration within the enormous and growing Python ecosystem and promoting seamless utilization for a range of computational and scripting tasks. With causal-learn, modification and extensions based on the existing implementation of causal discovery methods also become plausible for developers and researchers who may not be familiar with Java or R, which could significantly accelerate the progress in related fields by lowering the threshold of the integration of causality into various pipelines.

%% file: sections/design.tex
\section{Design}
The design philosophy of causal-learn is centered around building an open-source, modular, easily extensible and embeddable Python platform for learning causality from data and making use of causality for various purposes. Due to the different goals, assumptions, and techniques between causal learning and traditional learning tasks, newcomers to the field often find it hard to get a clear picture of the developments in modern causality research. Thus, we briefly introduce the algorithms and functionalities in causal-learn with a special focus on their use cases and suitable application scenarios.

\subsection{Search methods}
Causal-learn covers representative causal discovery methods across all major categories with official implementation of most algorithms. We briefly introduce the methods as follows. It is worth noting that we are actively updating the library to incorporate latest algorithms.
\begin{itemize}
    \item \textbf{Constraint-based causal discovery methods.} Current algorithms under that category are PC \citep{spirtes2000causation}, FCI \citep{spirtes1995causal}, and CD-NOD \citep{huang2020causal}. PC is a classical and widely-used algorithm with consistency guarantee under independent and identically distributed (i.i.d.) sampling assuming no latent confounders, the faithfulness assumption, and the causal Markov condition, which has been extensively applied in many fields. By continuously applying (conditional) independence tests on subsets of variables of increasing size in a smart way, its search procedure returns a Markov Equivalence Class (MEC), of which the graphical object consists of a mixture of directed and undirected edges, known as a Completed Partially Directed Acyclic Graph (CPDAG). PC is highly adaptable to various use cases, facilitated by the selection of an appropriate independence test; it can handle data with different assumptions, such as Fisher-Z test \citep{fisher1921014} for linear Gaussian data, Chi/G-squared test \citep{tsamardinos2006max} for discrete data, and Kernel-based Conditional Independence (KCI) test \citep{zhang2011kernel} for the nonparametric case. Moreover, causal-learn provides an extension, Missing-Value PC (MV-PC) \citep{tu2019causal}, to address issues of missing data. Furthermore, we have implemented FCI for causal structures that include hidden confounders (it indicates the possible existence of hidden confounders whenever the possibility cannot be excluded, but it cannot help determine possible relations among them), and causal discovery from nonstationary/heterogeneous data (CD-NOD). These constraint-based methods offer wide applicability as they can accommodate various types of data distributions and causal relations, provided that appropriate conditional independence testing methods are utilized. However, genenerally speaking, they may not be able to determine the complete causal graph uniquely and, accordingly, there usually exist some undirected edges in the returned CPDAGs.

    \item \textbf{Score-based causal discovery methods.} Different from the search style of constraint-bed methods, score-based methods find the causal structure by optimizing a properly defined score function. Greedy Equivalence Search (GES) \citep{chickering2002optimal} is a well-known two-phase procedure that directly searches over the space of equivalence classes. Similarly, exact search (e.g., A* \citep{yuan2013learning}, Dynamic Programming \citep{silander2006simple}), and permutation-based search (e.g., GRaSP \citep{lam2022greedy}) apply different search strategies to return a set of the sparsest Directed Acyclic Graphs (DAGs) that contains the true model under assumptions strictly weaker than faithfulness. These score-based methods are versatile, able to accommodate a wide array of data and causal relations by choosing suitable score functions, such as BIC \citep{schwarz1978estimating} for linear Guassian data, BDeu \citep{buntine1991theory} for discrete data, and Generalized Score \citep{huang2018generalized} for the nonparametric case. The choice of score function can be conveniently adjusted as a hyperparameter.

    \item \looseness=-1 \textbf{Causal discovery methods based on constrained functional causal models.} While constraint-based and score-based methods offer flexibility through the selection of an appropriate independence test or score function, they are limited to returning equivalence classes, yielding non-unique solutions where the causal direction between certain variable pairs remains indeterminate. In contrast, assuming specific Functional Causal Models (FCMs)--that is, functions in a particular functional class to specify how the effect is generated from its direct causes and noise--allows for the full determination of the causal structure, albeit at the cost of certain trade-offs. Causal-learn incorporates algorithms based on several FCM variants, capable of producing unique causal directions. Examples include the linear non-Gaussian acyclic model (LiNGAM) \citep{shimizu2006linear} and its variant, i.e., DirectLiNGAM \citep{shimizu2011directlingam}, which have been extensively applied for non-Gaussian noises with linear relations. VAR-LiNGAM \citep{hyvarinen2010estimation}, which combines LiNGAM with vector autoregressive models (VAR), to estimate both time-delayed and instantaneous causal relations from time series. RCD \citep{maeda2020rcd}, an extension of LiNGAM, allows for hidden confounders, while CAM-UV \citep{maeda2021causal} further extends this to the nonlinear additive noise case. In addition, the additive noise model (ANM) \citep{hoyer2008nonlinear} has been proven to be identifiable in the presence of nonlinearity and additive noises. Furthermore, we have also incorporated the post-nonlinear (PNL) causal model \citep{zhang2009identifiability}, a highly general form (with LiNGAM and ANM as special cases) that has been demonstrated to be identifiable in the generic case, barring five specific situations described in \citep{zhang2009identifiability}.

    \item \textbf{Causal representation learning: Finding causally related hidden variables.} Latent variables play an instrumental role in a multitude of real-world scenarios, often acting as hidden confounders that influence observed variables. Unfortunately, most existing methods may fail to produce convincing results in cases with latent variables (confounders). In causal-learn, we implement the Generalized Independent Noise (GIN) condition \citep{xie2020generalized} for estimating linear non-Gaussian latent variable causal model, which allows causal relationships between latent variables and multiple latent variables behind any two observed variables. This promises to improve the detection and understanding of the complex, often hidden, causal structures that govern real-world phenomena.

\end{itemize}

Besides, causal-learn also has Granger causality \citep{granger1969investigating, granger1980testing} implemented for statistical but not causal\footnote{As mentioned by Granger, Granger causality is not necessarily true causality. In fact, If one assumes 1) that there is no latent confounding process, 2) that the data are sampled at the right causal frequency, and 3) that there are no instantaneous causal influences, then Granger causality defined by Granger \citep{granger1980testing} can be seen as causal relations that can be discovered from stochastic processes with constraints-based methods such as PC. Of course, if those assumptions are violated, one may still apply Granger causal analysis, but the estimated relations may not be true causal influences.} time series analysis. Through the collective efforts of various teams and the contributions of the open-source community, causal-learn is always under active development to incorporate the most recent advancements in causal discovery and make them available to both practitioners and researchers.

\subsection{(Conditional) independence tests}

In addition to its comprehensive search methods, causal-learn also provides a variety of (conditional) independence tests as independent modules. Besides being an essential parts of several search methods, these tests can also be independently utilized and seamlessly integrated into existing statistical analysis pipelines. Currently,the library features a diverse array of such tests including Fisher-z test \citep{fisher1921014}, Missing-value Fisher-z test, Chi-Square test, Kernel-based conditional independence (KCI) test and independence test \citep{zhang2011kernel}, and G-Square test \citep{tsamardinos2006max}, each with distinct capabilities and benefits. The Fisher-z test is ideally suited for linear-Gaussian data, while the Missing-value Fisher-z test addresses the challenges of missing values by implementing a testwise-deletion approach. For categorical variables, the Chi-Square and G-Square tests are most effective. For users interested in a nonparametric test or the case with mixed categorical and continuous data types, the KCI test is an option. Overall, the range of tests offered by causal-learn underscores its versatility in handling diverse data types.

\subsection{Score functions}
\looseness=-1
Moreover, a diverse range of score functions is available in \textit{causal-learn}. These score functions quantify the goodness of fit of a model to the data, a crucial measure in score-based causal discovery methods, and can also be utilized independently for model selection in a broader range. Among these, the Bayesian Information Criterion (BIC) score \citep{schwarz1978estimating} is used extensively, offering a balance between model complexity and fit to the data. Another important score function is the Bayesian Dirichlet equivalent uniform (BDeu) score \citep{buntine1991theory}. This score function, especially beneficial for discrete data, incorporates a uniform prior over the set of Bayesian networks. Additionally, the Generalized Score \citep{huang2018generalized} is also available in causal-learn, which offers the flexibility to accommodate more complex scenarios and is beneficial for nonparametric cases where the true data-generating process does not align with the assumptions of BIC (linear Gaussian) or BDeu (discrete).

\subsection{Utilities}

Causal-learn further offers a suite of utilities designed to streamline the assembly of causal analysis pipelines. The package features a comprehensive range of graph operations encompassing transformations among various graphical objects integral to causal discovery. These include Directed Acyclic Graphs (DAGs), Completed Partially Directed Acyclic Graphs (CPDAGs), Partially Directed Acyclic Graphs (PDAGs), and Partially Ancestral Graphs (PAGs). Additionally, to enhance the convenience of experimental processes, \textit{causal-learn} features a set of commonly used evaluation metrics to appraise the quality of the causal graphs discovered. These metrics include precision and recall for arrow directions or adjacency matrices, along with the Structural Hamming Distance \citep{acid2003searching}.

\subsection{Demos, documentation, and benchmark datasets}

The \textit{causal-learn} package also contains extensive usage examples of all search methods, (conditional) independence tests, score functions, and utilities at 
\\ \centerline{ \url{https://github.com/py-why/causal-learn/tree/main/tests}.} 
\\
Furthermore, detailed documentation is available at \\
\centerline{\url{https://causal-learn.readthedocs.io/en/latest/}.} \\
It is worth noting that it also includes a collection of well-tested benchmark datasets--since ground-truth causal relations are often unknown for real data, evaluation of causal discovery methods has been notoriously known to be hard, and we hope the availability of such benchmark datasets can help alleviate this issue and inspire the collection of more real-world datasets with (at least partially) known causal relations.

%% file: sections/example.tex
\section{Example}

\begin{figure}[h]
    \centering
    \includegraphics[width=\linewidth]{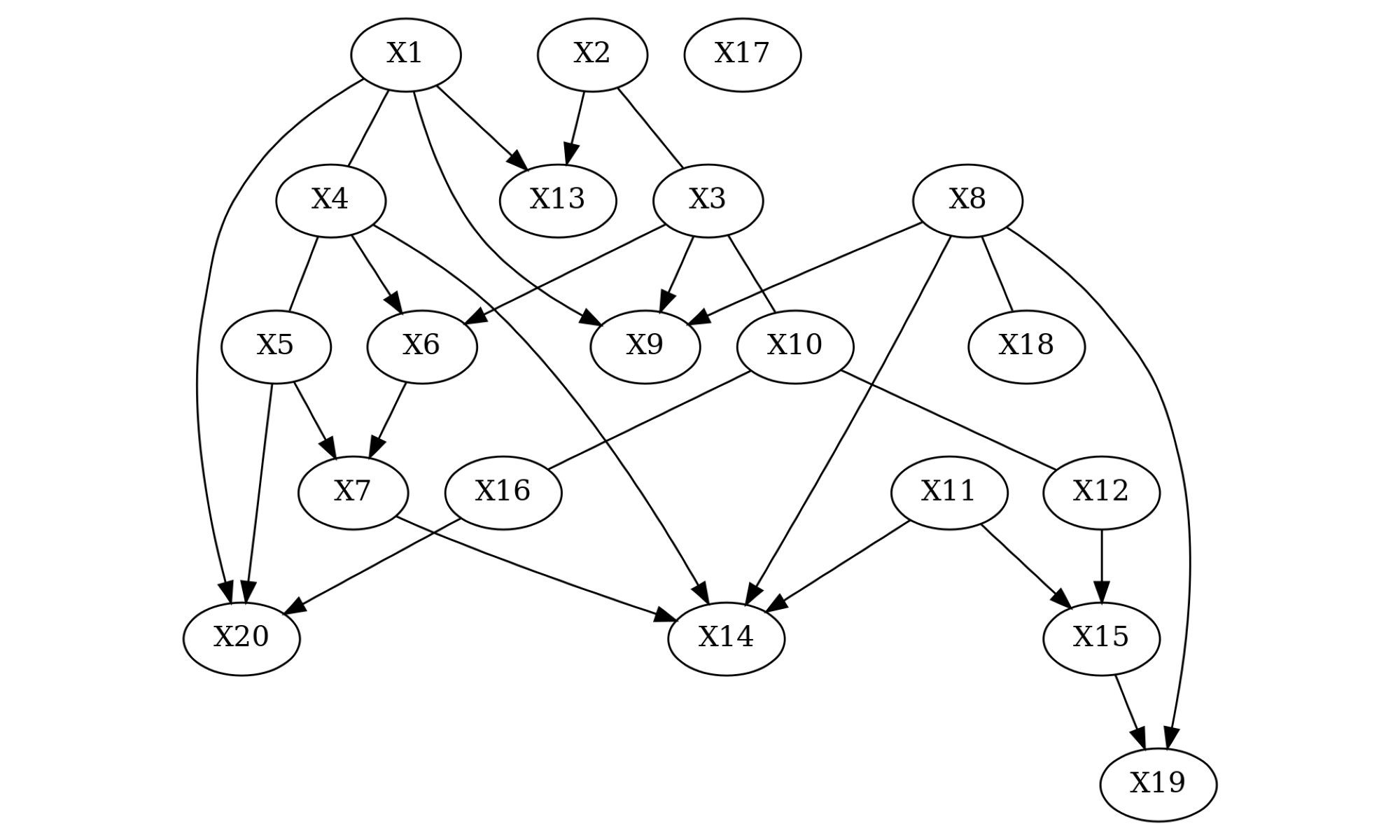}
    \caption{Visualization of the causal graph returned by \textit{causal-learn} with PC algorithm.}
    \label{fig:example}
\end{figure}

In this section, let us demonstrate how \textit{causal-learn} discovers causal relations from observational data in one line of code. First, we could easily install the library via pip:
\begin{lstlisting}
pip install causal-learn
\end{lstlisting}
Then we are ready to take a look into the causal world. Causal discovery in Python is as simple as follows:
\begin{lstlisting}
# apply PC with default parameters
cg = pc(data)

# visualization
cg.draw_pydot_graph()
\end{lstlisting}
The visualization of the returned causal graph is shown in Figure \ref{fig:example}.

\vspace{-1em}

%% file: sections/conclusion.tex
\section{Conclusion}

\looseness=-1
The \textit{causal-learn} library serves as a comprehensive toolset for causal discovery, significantly advancing the field of causal analysis and its applications in domains such as machine learning. It provides a robust platform for not only applying causal analysis techniques but also for facilitating the development of novel or enhanced algorithms. This is achieved by providing an infrastructure fully in Python that allows users to efficiently modify, extend, and tailor existing implementations, contribute new ones, and maintain high-quality standards. Given the current demand for causal learning and the rapid progress in this field, coupled with the active development and contribution from our team and the community, the \textit{causal-learn} library is poised to bring causality into an indispensable component across diverse disciplines.